\documentclass[]{wechat}

\usepackage{amsmath,amsfonts,bm}

\def\eqref#1{equation~\ref{#1}}

\def\1{\bm{1}}

\DeclareMathAlphabet{\mathsfit}{\encodingdefault}{\sfdefault}{m}{sl}
\SetMathAlphabet{\mathsfit}{bold}{\encodingdefault}{\sfdefault}{bx}{n}

\usepackage{url}
\usepackage{threeparttable}
\usepackage[table]{xcolor} 
\usepackage[bottom]{footmisc} 

\usepackage{CJKutf8}

\usepackage{amsmath,amssymb}
\usepackage{booktabs}
\usepackage{tabularx}
\usepackage{array}
\usepackage{graphicx}
\usepackage{xcolor}
\usepackage{bbm}
\graphicspath{{figure/figs/}}

\newcolumntype{Y}{>{\centering\arraybackslash}X}

\newcommand{\linkicon}[1]{\raisebox{-0.2em}{\includegraphics[height=1em]{#1}}}
\newcommand{\hficon}{\linkicon{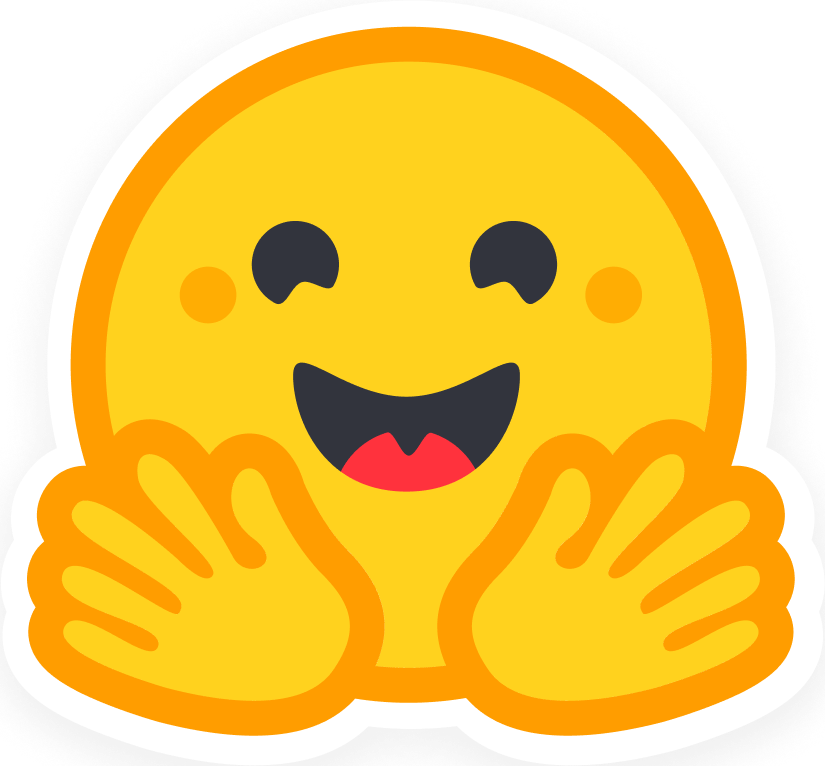}}
\newcommand{\ghicon}{\linkicon{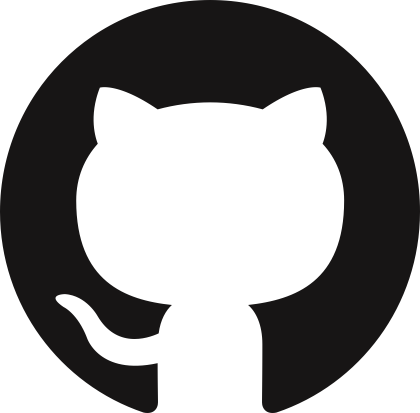}}
\newcommand{\hflink}{https://huggingface.co/collections/tencent/wemm-embedding}
\newcommand{\ghlink}{https://github.com/Tencent/WeMM-Embedding}
\newcommand{\resourcelinks}{%
  \begingroup
  \small\raggedright
  \setlength{\parindent}{0pt}%
  \setlength{\parskip}{0pt}%
  \href{\hflink}{\makebox[1.6em][l]{\hficon}\texttt{\hflink}}\par
  \href{\ghlink}{\makebox[1.6em][l]{\ghicon}\texttt{\ghlink}}\par
  \endgroup
}

\title{WeMM-Embedding: WeChat Multi-Modal Embedding Technical Report} 
\author{%
\fontsize{11.2pt}{13.2pt}\selectfont
\centerline{%
   Junjie Zhou$^{*}$, Ke Mei$^{*\dagger}$, Lei Li$^{*}$,
   Tianyi Wang, Fengyun Rao$^{\ddagger}$, Jing LYU
}
\vspace{6pt}
\centerline{%
   WeChat Vision, Tencent Inc.
}
\vspace{-24pt}
}

\vspace{-15pt}
\abstract{
Universal multimodal embeddings are becoming a core component of modern AI systems, enabling heterogeneous content to be represented in a shared space for applications such as retrieval, recommendation, classification, and agentic systems. In this report, we present \textbf{\textit{WeMM-Embedding}}, a family of universal multimodal embedding models supporting text, images, videos, visual documents, and arbitrarily interleaved multimodal inputs with flexible output dimensions. The family comprises 2B, 4B, and 9B variants and is trained in two stages: a large-scale multimodal alignment stage, followed by a refinement stage using curated data, fine-grained relevance supervision, and cross-scale knowledge transfer. Across extensive evaluations, WeMM-Embedding achieves leading performance on multiple public benchmarks. Notably, the 2B variant already surpasses the previously leading 8B open-source baseline on MMEB-v2, while the 9B variant further achieves a new state-of-the-art overall score of 80.6. WeMM-Embedding also demonstrates strong practical performance across WeChat applications, with substantial gains on a 26-task in-house benchmark and consistent improvements across 14 online A/B tests. It has been deployed at scale across recommendation and search applications, including WeChat Channels, Official Accounts, Moments, and e-commerce services. We have released the model weights and code to facilitate future research.
\par\vspace{5pt}
\resourcelinks
\vspace{-3.5pt}
}
\date{\today}

\begin{document}
\maketitle

\begingroup
\renewcommand{\thefootnote}{}
\footnotetext{\scriptsize
$^{*}$Core contributors.
\quad
$^{\dagger}$Project leader.
\quad
$^{\ddagger}$Corresponding author: \url{fengyunrao@tencent.com}
}
\endgroup

\vspace{-22pt}

\begin{figure}
\centering
\includegraphics[width=\textwidth]{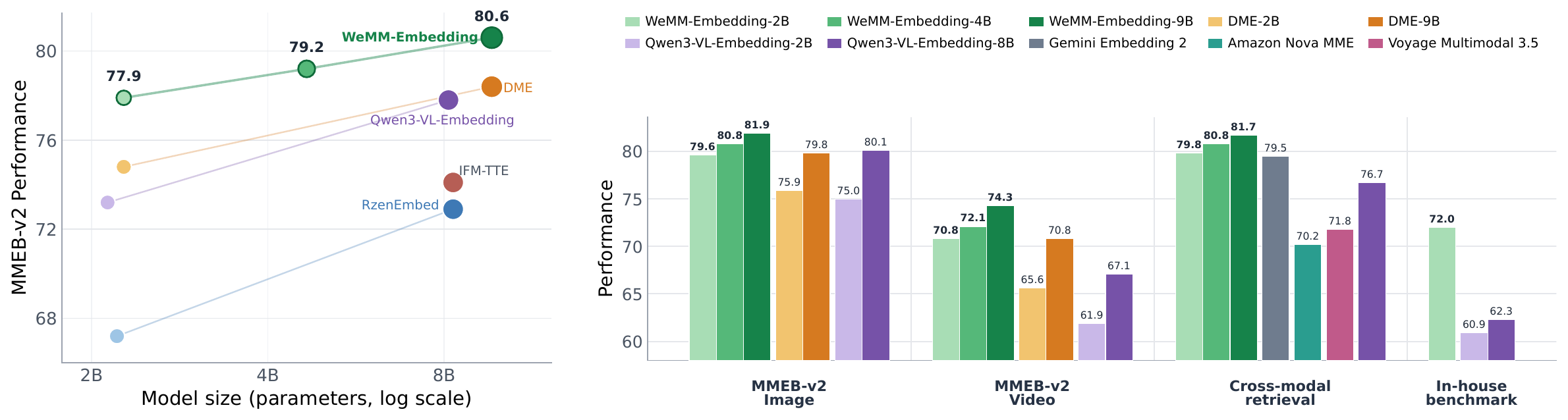}
\caption{\textbf{Performance and efficiency overview of WeMM-Embedding.} \textbf{Left:} MMEB-v2 overall performance across different model sizes, compared with representative baselines. \textbf{Right:} Aggregate performance on the MMEB-v2 image and video subsets, a 12-dataset cross-modal retrieval suite, and the 26-task in-house benchmark.}
\label{fig:performance_overview}
\end{figure}

\section{Introduction}
\label{sec:introduction}

Multimodal embedding models have become fundamental components of modern AI systems, owing to their ability to map heterogeneous inputs, such as text, images, and videos, into a shared dense representation space. Such representations support a broad range of downstream applications, including classification~\citep{deng2009imagenet,kay2017kinetics}, any-to-any retrieval~\citep{cirr-liu2021image,chang2022webqa,yuan2025momentseeker}, clustering~\citep{xiao2025mieb}, recommendation~\citep{DBLP:journals/corr/abs-2502-18965,chen2025onesearch}, and agentic systems~\citep{hu2024mrag,xu2026mem,chhikara2025mem0}. Early CLIP-style models~\citep{clip-radford2021learning,siglip2023,evaclip-sun2023eva} established the effectiveness of large-scale cross-modal alignment through modality-specific dual encoders. However, their modality-specific encoding pathways do not naturally support the joint representation of inputs that combine multiple modalities, such as interleaved text--image documents, composed multimodal queries, or videos paired with transcripts. Subsequent work incorporated pretrained visual encoders as visual tokenizers for text encoders, extending this paradigm to broader any-to-any matching~\citep{vista,marvel-zhou2023unlock}. Nevertheless, these encoder-based approaches remain limited in generality as multimodal embedding expands toward increasingly diverse tasks and input compositions.

The emergence of Multimodal Large Language Models (MLLMs)~\citep{gpt4,llava-liu2023visual,internvl2} has driven rapid progress in universal multimodal embedding models. MLLMs naturally support arbitrary interleaved combinations of text, images, and videos, while their broad pretrained capabilities provide a strong foundation for representation learning across diverse tasks. Motivated by these advantages, early studies explored adapting the hidden states of MLLMs into general-purpose embeddings and demonstrated the feasibility of continued contrastive training on paired multimodal data~\citep{nv-mm-embed2024,vlm2vec2024}. More recent advances in large-scale paired multimodal data synthesis and knowledge distillation have further improved model performance, making MLLM-based embedding an increasingly prominent paradigm for universal multimodal representation learning~\citep{zhang2024gme,zhou2025megapairs,chen2025mme5,qwen3vl-emb}.

In this work, we introduce \textbf{WeMM-Embedding}, a family of universal multimodal embedding models spanning 2B, 4B, and 9B scales. To build a strong and general-purpose multimodal embedding family, we devise a two-stage training strategy that progressively moves from broad multimodal alignment to finer-grained relevance learning. In the first stage, the models are trained on several hundred million heterogeneous pairs spanning diverse modalities, tasks, and domains, establishing broad multimodal coverage and a strong initial representation space. In the second stage, we further refine the models on a carefully curated corpus with improved semantic balance, higher data quality, and more challenging negatives, while introducing richer relevance supervision and cross-scale knowledge transfer to strengthen fine-grained matching. This progressive training strategy allows WeMM-Embedding to retain broad capability coverage while continuously improving relevance modeling and representation quality.

WeMM-Embedding achieves state-of-the-art performance across a broad range of public benchmarks and shows strong practical performance in real-world applications. As summarized in Figure~\ref{fig:performance_overview}, the compact 2B variant already surpasses previously leading 8B open-source baselines on MMEB-v2~\cite{vlm2vec-v2}, highlighting the strong parameter efficiency of WeMM-Embedding. Scaling to 9B further raises the overall score to 80.6, ranking first on the official MMEB-v2 leaderboard and outperforming all listed open-source and proprietary models.\footnote{Leaderboard status as of August 24, 2026, according to the \href{https://huggingface.co/spaces/TIGER-Lab/MMEB-Leaderboard}{official MMEB leaderboard}.} On the cross-modal retrieval suite reported in Gemini Embedding 2~\cite{gemini-emb-2}, the 2B model also compares favorably with leading proprietary models. Beyond public benchmark evaluations, WeMM-Embedding also demonstrates strong practical performance, with substantial gains on a 26-task in-house benchmark and consistent improvements across 14 online A/B tests across WeChat applications. It has also been deployed at scale across recommendation and search applications, including WeChat Channels, Official Accounts, Moments, and e-commerce services. Together, these results establish a new performance--efficiency frontier for universal multimodal embedding models and demonstrate strong practical value in large-scale deployment.

\section{Data Construction}
\label{sec:data}

To support general-purpose multimodal representation learning, we collect and synthesize several hundred million training examples spanning diverse modalities, domains, and tasks. We express these heterogeneous examples in a unified pair-based format, allowing them to be incorporated into a common training framework while preserving their original supervision signals. We further construct a curated collection that complements the large-scale data with greater semantic diversity, higher data quality, and more informative supervision.

\subsection{Large-Scale Training Data Collection}
\label{sec:data_collection}

Our training data are collected from public datasets and web-scale weakly supervised sources, and are further supplemented with task-oriented synthetic data and in-house collections. The resulting data span text, images, videos, and their interleaved combinations across diverse domains and task settings.

\paragraph{Unified Pair-Based Format.}

Despite substantial variation in input structures and supervision signals, these heterogeneous tasks can all be formulated as matching a source instance against one or more target candidates. Accordingly, we represent each training example in a unified pair-based format:

\begin{equation}
    z_i
    =
    \left(
        I_i,\,
        q_i,\,
        c_i,\,
        \mathcal{N}_i,\,
        y_i
    \right),
    \label{eq:pair_data_format}
\end{equation}

where $I_i$ denotes an optional task-specific instruction, $q_i$ denotes the source instance, $c_i$ denotes the paired target, $\mathcal{N}_i$ denotes an optional set of explicit hard negatives, and $y_i$ denotes an optional graded relevance score. Both $q_i$ and $c_i$ may contain text, images, videos, or interleaved combinations of these modalities. Each element of $\mathcal{N}_i$ is drawn from the same target candidate space as $c_i$ and has the corresponding target-side modality structure. When provided, $I_i$ specifies the intended matching relation and helps distinguish tasks with similar input modalities. For standard source--target pairs, $c_i$ serves as the positive target, while both $\mathcal{N}_i$ and $y_i$ may be omitted. When explicit hard negatives are available, $\mathcal{N}_i$ contains candidates that are related to $q_i$ but less relevant than $c_i$. For datasets with graded relevance annotations, $y_i$ retains the original ordinal or continuous relevance score assigned to the pair $(q_i,c_i)$ and is subsequently used to construct relative ordering constraints during training.

Under this formulation, examples from different tasks share a common source--target structure and can be organized into task-specific batches within a unified multi-task training pipeline. For standard pairwise data, targets from other examples in the same batch serve as in-batch negatives. For tasks with shared target spaces, such as classification, repeated targets may introduce false negatives; these collisions are excluded through semantic target masking, as detailed in Section~\ref{sec:training_strategy}.

\begin{figure}[t]
\centering
\includegraphics[width=\linewidth]{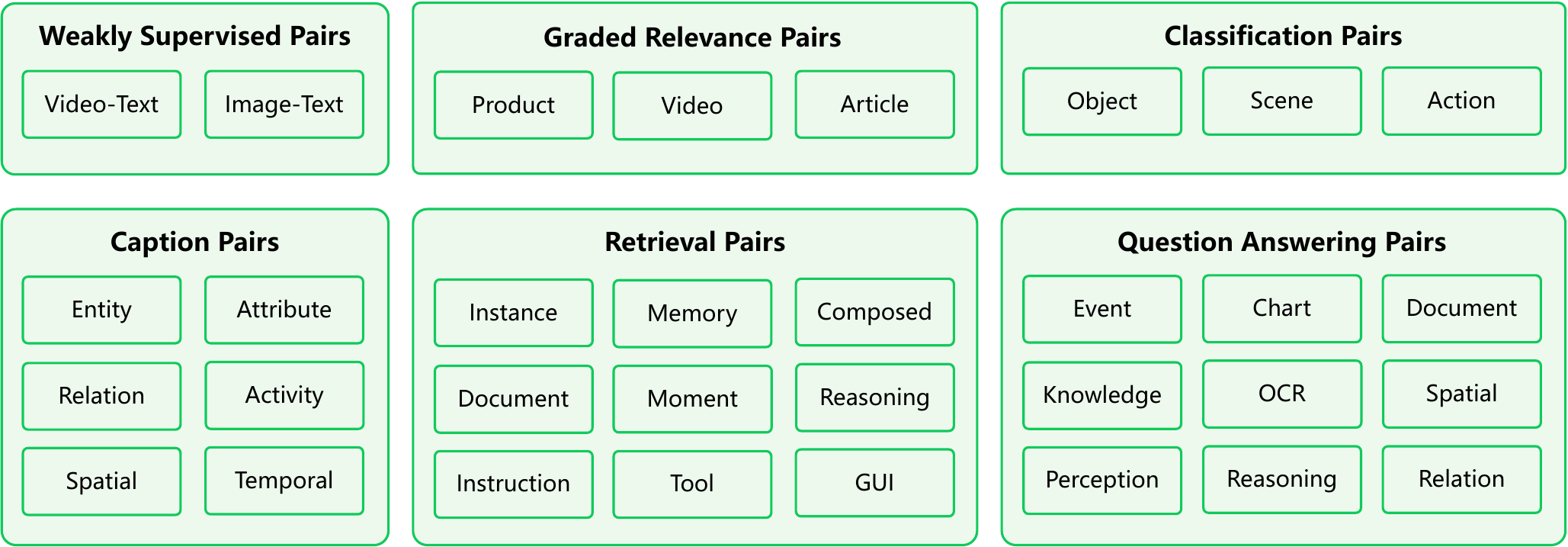}
\caption{\textbf{Overview of our multimodal training data.} Major data families and representative coverage across diverse task settings and content domains.}
\label{fig:data_overview}
\end{figure}

\paragraph{Data Coverage and Composition.}

As shown in Figure~\ref{fig:data_overview}, our training data cover diverse forms of supervision, task settings, and content domains. The training corpus includes several major types of data and supervision, with representative examples illustrated in the figure. We describe the main components below:

\begin{itemize}
    \item \textbf{Weakly Supervised Pairs.} We collect large-scale image--text and video--text pairs from public datasets and web-scale weakly supervised sources, where visual and textual content are associated through naturally occurring correspondence rather than explicit visual descriptions. These pairs provide broad but relatively coarse supervision across diverse visual content. 

    \item \textbf{Caption Pairs.} We pair images and videos with captions that explicitly describe their visual content, ranging from concise summaries to detailed descriptions of entities, attributes, relations, spatial context, actions, and events. Compared with weakly supervised pairs, these examples provide more direct and fine-grained visual--language correspondence.

    \item \textbf{Retrieval Pairs.} Retrieval examples associate queries with relevant candidates across text, images, videos, and interleaved multimodal inputs. They range from conventional unimodal and cross-modal matching to more challenging settings involving composed queries, reasoning, instructions, long contexts, spatial grounding, temporal localization, and agent-related retrieval.

    \item \textbf{Classification Pairs.} We reformulate classification datasets as source--label pairs, where the target is represented by either a class name or a natural-language description of the corresponding category. These examples cover a variety of image- and video-based recognition tasks, including object, scene, and action classification.

    \item \textbf{Multimodal Question-Answer Pairs.} We pair textual or multimodal questions with answer targets across a range of capabilities, including visual perception, relational and spatial understanding, optical character recognition, knowledge-intensive understanding, reasoning, document and chart comprehension, and event understanding.

    \item \textbf{Graded Relevance Pairs.} In the large-scale collection, graded supervision takes the form of manually assigned discrete relevance levels for source--target pairs. These labels distinguish multiple degrees of relevance beyond binary matching and support ranking-oriented training~\citep{graded-level}. Reranker-derived relevance scores are introduced separately during the subsequent data curation stage.
\end{itemize}

\subsection{Curated Data Construction}
\label{sec:data_refinement}

Alongside the large-scale collection, we construct a curated dataset approximately one tenth its size. The curated dataset focuses on improving semantic balance and data quality while introducing more informative supervision. Its construction includes Semantic-ID-guided resampling, quality control, and selective hard-negative enrichment.

\paragraph{Semantic-ID-Guided Resampling.}

Although the large-scale collection covers a broad range of domains and tasks, its semantic distribution remains skewed toward frequent patterns. Inspired by Semantic IDs~\citep{tiger}, we derive a discrete identifier for each source--target pair to characterize this distribution and guide resampling. For each pair, the side with the longer serialized token sequence is encoded using an intermediate checkpoint of WeMM-Embedding. We then fit a three-level residual k-means quantizer (RQ-KMeans)~\citep{rq-kmeans} to the resulting representations. Each representation is subsequently mapped through the learned codebooks to obtain a three-element Semantic ID.
We use the assignment density at each codebook level to guide resampling. Examples associated with densely populated codes are sampled at lower rates, whereas those mapped to less populated codes are retained at higher rates. This reduces repeated exposure to frequent semantic patterns without enforcing a uniform distribution over Semantic IDs.

\paragraph{Data Quality Refinement.}

The resampled training examples are further refined using a multimodal large language model. Based on the corresponding task context, the model assesses whether each source--target pair reflects the intended matching relation and filters out mismatched examples. We also refine the textual fields where necessary. For example, weakly supervised image--text and video--text pairs offer broad coverage but may contain noisy or factually inaccurate descriptions; in such cases, the model corrects factual inconsistencies while preserving the original alt-text style and level of detail.

\paragraph{Hard-Negative Construction.}

A subset of the refined training examples is further enriched with explicit hard negatives to improve discrimination among semantically similar candidates. The construction procedure varies with the target modality. For text targets, multimodal large language models generate plausible but incorrect candidates based on the source and positive target. For image and video targets, intermediate checkpoints of our WeMM-Embedding models are used to retrieve semantically similar candidates from task-specific candidate pools. For a smaller subset of the mined candidate sets, reranking models are further used to assign relevance scores, providing finer-grained supervision among difficult candidates. The corresponding training objective is introduced in Section~\ref{sec:training_strategy}.

\section{Modeling and Training Strategy}
\label{sec:training}

WeMM-Embedding comprises three universal multimodal embedding models with 2B, 4B, and 9B parameters, built on the corresponding natively multimodal Qwen3.5 backbones~\citep{qwen35blog}. This section first introduces how heterogeneous multimodal inputs are encoded into dense representations, and then presents the two-stage strategy used to train the model family.

\subsection{Modeling}
\label{sec:embedding_formulation}

WeMM-Embedding is built upon the Qwen3.5 architecture. Benefiting from its native support for heterogeneous and interleaved multimodal inputs, WeMM-Embedding can encode arbitrary combinations of text, images, and videos into unified dense representations. We adopt last-token pooling by appending a dedicated \texttt{<embedding>} token to the input sequence and using its final-layer hidden state as the output representation.

Formally, an input instance may contain an optional task-specific instruction together with an ordered sequence of multimodal segments:

\begin{equation}
    \mathcal{D}
    =
    \left\langle
        I_{\mathrm{inst}},
        x_1,
        x_2,
        \ldots,
        x_m
    \right\rangle,
    \label{eq:multimodal_input}
\end{equation}

where $I_{\mathrm{inst}}$ denotes the optional instruction and each $x_i$ may correspond to text, an image, or a video. Textual segments are converted into token embeddings through the native tokenizer and embedding layer, while visual segments are converted into visual token representations through the native visual processing pipeline. The textual and visual tokens are arranged according to the original order of the input segments, followed by a dedicated \texttt{<embedding>} token:

\begin{equation}
    \mathbf{S}
    =
    \left[
        \mathbf{z}_1,
        \mathbf{z}_2,
        \ldots,
        \mathbf{z}_N,
        \mathbf{z}_{\mathrm{emb}}
    \right],
    \label{eq:multimodal_sequence}
\end{equation}

where each $\mathbf{z}_i$ denotes a textual or visual token embedding, and $\mathbf{z}_{\mathrm{emb}}$ denotes the token embedding of \texttt{<embedding>}. The resulting multimodal sequence is then processed by the LLM backbone $G_{\theta}$:

\begin{equation}
    \mathbf{H}
    =
    G_{\theta}(\mathbf{S})
    =
    \left[
        \mathbf{h}_1,
        \mathbf{h}_2,
        \ldots,
        \mathbf{h}_N,
        \mathbf{h}_{\mathrm{emb}}
    \right].
    \label{eq:multimodal_hidden_states}
\end{equation}

By default, the \texttt{<embedding>} token is placed at the end of the sequence and attends to all preceding textual and visual content under the native causal attention mask. The same causal formulation also allows multiple \texttt{<embedding>} tokens to be inserted at different sequence positions. For example, when a video is followed by its automatic speech recognition (ASR) transcript, placing one token after the video tokens and another at the end of the sequence enables video-only and joint video--text representations to be extracted within a single forward pass, supporting downstream applications with different modality requirements.

The final-layer hidden state corresponding to each \texttt{<embedding>} token is L2-normalized to obtain the output representation:

\begin{equation}
    \mathbf{e}_{\mathcal{D}}
    =
    \frac{
        \mathbf{h}_{\mathrm{emb}}
    }{
        \left\|
            \mathbf{h}_{\mathrm{emb}}
        \right\|_2
    }.
    \label{eq:embedding_representation}
\end{equation}

In addition, WeMM-Embedding supports flexible embedding dimensions through Matryoshka Representation Learning (MRL)~\citep{MRL2022}. Given the final hidden state $\mathbf{h}_{\mathrm{emb}}\in\mathbb{R}^{D}$, an embedding of dimension $d\leq D$ is obtained by retaining its first $d$ dimensions and applying L2 normalization:

\begin{equation}
    \mathbf{e}_{\mathcal{D}}^{(d)}
    =
    \frac{
        \mathbf{h}_{\mathrm{emb},1:d}
    }{
        \left\|
            \mathbf{h}_{\mathrm{emb},1:d}
        \right\|_2
    },
    \qquad
    d\in\mathcal{D}_{\mathrm{MRL}},
    \label{eq:mrl_representation}
\end{equation}

where $\mathcal{D}_{\mathrm{MRL}}$ denotes the predefined set of supported embedding dimensions. At inference time, embeddings at all supported dimensions can be obtained from a single forward pass through prefix truncation and re-normalization.

\subsection{Training Strategy}
\label{sec:training_strategy}

We train WeMM-Embedding using a two-stage strategy. The first stage establishes a general multimodal embedding space through large-scale multi-task alignment. The second stage continues training on curated data, combining contrastive learning with explicit hard negatives, selective reranker supervision, and embedding distillation from a larger model. Both stages operate on the unified pair-based representation introduced in Section~\ref{sec:data_collection}, while each batch uses the objective supported by its supervision signals.

\subsubsection{Stage 1: Large-Scale Multimodal Alignment}
\label{sec:stage1_training}

In the first stage, we train WeMM-Embedding on the large-scale collection described in Section~\ref{sec:data_collection}, comprising several hundred million source--target pairs across diverse modalities, domains, and tasks. Each batch is constructed from a consistent data source, while batches from different tasks are interleaved throughout training. Standard paired examples are optimized with contrastive learning~\cite{infonce}, whereas examples carrying native graded relevance annotations use a score-gap-weighted CoSENT-style ranking objective~\cite{cosent}.

\paragraph{Contrastive Learning.}

For standard paired data, we optimize source--target alignment using an InfoNCE objective~\citep{infonce}. Each batch is drawn from a single dataset, keeping the task definition and candidate space consistent within the batch. For each source, its paired target is treated as the positive, while targets associated with other examples serve as in-batch negatives. When explicit hard negatives are available, they are incorporated into the same negative pool. This construction provides informative task-consistent negatives, but may also introduce collisions when different examples contain nearly identical sources or targets. We therefore apply duplicate-aware masking on both sides of the pair before computing the loss.

Let
$\mathcal{C}_j=\{c_j^{+}\}\cup\mathcal{N}_j$
denote the candidates associated with source $q_j$, where $\mathcal{N}_j$ is empty for ordinary paired examples. Given a batch of $B$ source--target pairs, the contrastive objective is defined as

\begin{equation}
    \mathcal{L}_{\mathrm{CL}}
    =
    -
    \frac{1}{B}
    \sum_{i=1}^{B}
    \log
    \frac{
        \exp\left(s(q_i,c_i^{+})/\tau\right)
    }{
        \exp\left(s(q_i,c_i^{+})/\tau\right)
        +
        \displaystyle
        \sum_{j=1}^{B}
        \sum_{c\in\mathcal{C}_j\setminus\{c_i^{+}\}}
        M_{i,j,c}
        \exp\left(s(q_i,c)/\tau\right)
    },
    \label{eq:contrastive_loss}
\end{equation}

where $s(\cdot,\cdot)$ denotes cosine similarity between normalized embeddings and $\tau$ is a learnable temperature parameter. The duplicate-aware mask is defined as

\begin{equation}
    M_{i,j,c}
    =
    \begin{cases}
        0,
        &
        \begin{aligned}
        &\text{if } j\neq i
        \text{ and }
        s(q_i,q_j)>\tau_{\mathrm{dup}},\\[-2pt]
        &\text{or if }
        s(c_i^{+},c)>\tau_{\mathrm{dup}},
        \end{aligned}
        \\[8pt]
        1,
        & \text{otherwise},
    \end{cases}
    \label{eq:duplicate_mask}
\end{equation}

where $\tau_{\mathrm{dup}}$ is the similarity threshold used to identify near-duplicate representations. Source-side masking excludes all candidates associated with a near-duplicate source, while target-side masking excludes candidates that closely match the current positive target. The latter applies to both in-batch targets and explicit hard negatives.

\paragraph{Graded Relevance Learning.}

Within the large-scale collection, a subset of source--target pairs is annotated with manually assigned discrete relevance levels. Such annotations are particularly useful for item-to-item retrieval and recommendation-oriented scenarios, where different pairs may exhibit varying degrees of relatedness. Treating these examples as ordinary positive pairs under the contrastive objective would ignore this graded structure. We therefore adopt a score-gap-weighted CoSENT-style objective~\citep{cosent}, which optimizes the relative ordering of pairwise similarities and assigns greater importance to comparisons with larger relevance gaps.

Formally, given a graded-relevance batch
\begin{equation}
    \mathcal{B}_{\mathrm{rel}}
    =
    \left\{
        (q_i,c_i,y_i)
    \right\}_{i=1}^{G},
    \label{eq:relevance_batch}
\end{equation}

let $s_i=s(q_i,c_i)$ denote the predicted similarity of the $i$-th source--target pair. For two examples $i$ and $j$, we define the relevance-gap weight as

\begin{equation}
    w_{ij}
    =
    \max
    \left(
        \left|y_i-y_j\right|,
        \epsilon
    \right),
    \label{eq:relevance_gap_weight}
\end{equation}

where $\epsilon$ is a small positive constant. The ranking loss associated with example $i$ is

\begin{align}
    \mathcal{L}_{i}^{\mathrm{Rel}}
    =
    \log
    \Bigg[
        1
        &+
        \sum_{j:y_i>y_j}
        w_{ij}
        \exp
        \left(
            \gamma
            \left[
                s_j-s_i
            \right]
        \right)
        \nonumber\\
        &+
        \sum_{j:y_j>y_i}
        w_{ij}
        \exp
        \left(
            \gamma
            \left[
                s_i-s_j
            \right]
        \right)
    \Bigg],
    \label{eq:relevance_anchor_loss}
\end{align}

and the batch objective is

\begin{equation}
    \mathcal{L}_{\mathrm{Rel}}
    =
    \frac{1}{G}
    \sum_{i=1}^{G}
    \mathcal{L}_{i}^{\mathrm{Rel}}.
    \label{eq:relevance_ranking_loss}
\end{equation}

This objective encourages pairs with higher relevance levels to receive higher similarities and gives greater weight to comparisons with larger label gaps. Pairs with equal labels impose no ordering constraint.

\paragraph{Matryoshka Representation Learning.}

We apply Matryoshka Representation Learning~\cite{MRL2022} to both contrastive and graded-relevance training. For each batch, the objective associated with its supervision is evaluated independently at every supported embedding dimension:

\begin{equation}
    \mathcal{L}_{X}^{\mathrm{MRL}}
    =
    \sum_{d\in\mathcal{D}_{\mathrm{MRL}}}
    \alpha_d
    \mathcal{L}_{X}^{(d)},
    \qquad
    X\in
    \left\{
        \mathrm{CL},
        \mathrm{Rel}
    \right\},
    \label{eq:stage1_mrl_objective}
\end{equation}

where $\mathcal{L}_{X}^{(d)}$ is computed using embeddings independently truncated and normalized at dimension $d$, and $\alpha_d$ denotes the corresponding loss weight. Thus, each interleaved batch is optimized using the multi-dimensional form of its corresponding task objective.

\subsubsection{Stage 2: Curated Fine-Tuning and Distillation}
\label{sec:stage2_training}

Following large-scale multimodal alignment, WeMM-Embedding is further trained on the curated dataset described in Section~\ref{sec:data_refinement}. The contrastive and graded-relevance objectives remain in use, while additional supervision is provided by reranking models and a larger embedding teacher. Reranking models provide query-specific ordering signals over mined candidates for selected tasks, whereas the embedding teacher transfers the similarity structure induced within each training batch.
\paragraph{Reranker Supervision.}

We train dedicated multimodal rerankers to provide fine-grained ordering supervision over query-specific candidate sets, each containing an annotated positive target and mined hard negatives. Building on the score-gap-weighted CoSENT formulation used for graded relevance learning in Section~\ref{sec:stage1_training}, we replace the manually assigned relevance levels with reranker scores and construct comparisons only among candidates associated with the same query. Specifically, for source $q_b$ with candidate set $\mathcal{C}_b=\{c_{b,1},\ldots,c_{b,K}\}$, the reranker scores $\hat{y}_{b,k}$ induce the ordered pair set

\begin{equation}
    \widehat{\mathcal{P}}_b
    =
    \left\{
        (i,j)
        \mid
        \hat{y}_{b,i}>\hat{y}_{b,j}
    \right\}.
    \label{eq:reranker_ordering}
\end{equation}

The corresponding objective is

\begin{equation}
    \mathcal{L}_{\mathrm{Rank}}
    =
    \frac{1}{B}
    \sum_{b=1}^{B}
    \log
    \left[
        1+
        \sum_{(i,j)\in\widehat{\mathcal{P}}_b}
        \omega_{b,ij}
        \exp
        \left(
            \gamma
            \left[
                s(q_b,c_{b,j})
                -
                s(q_b,c_{b,i})
            \right]
        \right)
    \right],
    \label{eq:reranker_ranking_loss}
\end{equation}

where
$\omega_{b,ij}=\max\!\left(\left|\hat{y}_{b,i}-\hat{y}_{b,j}\right|,\epsilon\right)$
weights each comparison according to the reranker score gap. Reranker-scored batches use $\mathcal{L}_{\mathrm{Rank}}$ in place of the standard contrastive objective.

Empirically, we find that reranking candidates retrieved by our embedding models does not consistently improve performance across multimodal tasks. Similar to observations reported in prior work~\citep{qwen3vl-emb}, stable gains are observed only on a limited subset of tasks. We therefore restrict reranker supervision to settings where it yields reliable improvements.

\paragraph{Embedding Distillation.}

To obtain teacher supervision with broader coverage, we additionally distill from a larger model in the same WeMM-Embedding family. Unlike reranker supervision, which relies on task-specific models and preconstructed candidate sets, embedding distillation derives online soft targets from the batch-wise source--target similarity distributions produced by the teacher. It therefore requires no additional offline relevance annotations and can be applied across heterogeneous batch types.

Formally, for a batch containing $B$ sources and a target pool of size $K$, let $\mathbf{A}^{T},\mathbf{A}^{S}\in\mathbb{R}^{B\times K}$ denote the source-to-target similarity matrices produced by the teacher and student:

\begin{equation}
    A^{T}_{ij}
    =
    \frac{s_T(q_i,c_j)}{\tau_T},
    \qquad
    A^{S}_{ij}
    =
    \frac{s_S(q_i,c_j)}{\tau_S},
    \label{eq:source_target_similarity}
\end{equation}

where $\tau_T$ and $\tau_S$ are the teacher and student temperatures. We further construct the reverse similarity matrices between the positive targets and the sources:

\begin{equation}
    \overline{A}^{T}_{ij}
    =
    \frac{s_T(c_i^{+},q_j)}{\tau_T},
    \qquad
    \overline{A}^{S}_{ij}
    =
    \frac{s_S(c_i^{+},q_j)}{\tau_S}.
    \label{eq:target_source_similarity}
\end{equation}

The corresponding row-wise relation distributions are

\begin{align}
    \mathbf{P}_{T,i}^{q\rightarrow c}
    &=
    \operatorname{Softmax}
    \left(
        \mathbf{A}^{T}_{i,:}
    \right),
    &
    \mathbf{P}_{S,i}^{q\rightarrow c}
    &=
    \operatorname{Softmax}
    \left(
        \mathbf{A}^{S}_{i,:}
    \right),
    \\
    \mathbf{P}_{T,i}^{c\rightarrow q}
    &=
    \operatorname{Softmax}
    \left(
        \overline{\mathbf{A}}^{T}_{i,:}
    \right),
    &
    \mathbf{P}_{S,i}^{c\rightarrow q}
    &=
    \operatorname{Softmax}
    \left(
        \overline{\mathbf{A}}^{S}_{i,:}
    \right).
\end{align}

We define the bidirectional embedding-distillation objective as

\begin{equation}
    \mathcal{L}_{\mathrm{Emb}}
    =
    \frac{1}{2B}
    \sum_{i=1}^{B}
    \left[
        D_{\mathrm{KL}}
        \left(
            \mathbf{P}_{T,i}^{q\rightarrow c}
            \,\Vert\,
            \mathbf{P}_{S,i}^{q\rightarrow c}
        \right)
        +
        D_{\mathrm{KL}}
        \left(
            \mathbf{P}_{T,i}^{c\rightarrow q}
            \,\Vert\,
            \mathbf{P}_{S,i}^{c\rightarrow q}
        \right)
    \right].
    \label{eq:embedding_distillation}
\end{equation}

The two terms align the teacher and student similarity distributions in the source-to-target and target-to-source directions, respectively. Unlike the one-hot supervision used in standard contrastive learning, the teacher distributions retain relative similarity differences among candidates, providing softer and more structured targets for transferring semantic relations. Empirically, this supervision is particularly valuable for our compact WeMM-Embedding variants and contributes substantially to their performance gains.

\paragraph{Training Configuration.}

For each Stage~2 batch, the task objective is selected according to its available supervision:

\begin{equation}
    \mathcal{L}_{\mathrm{Task}}
    =
    \begin{cases}
        \mathcal{L}_{\mathrm{CL}}^{\mathrm{MRL}},
        & \text{for standard paired or hard-negative batches}, \\[3pt]
        \mathcal{L}_{\mathrm{Rel}}^{\mathrm{MRL}},
        & \text{for graded-relevance batches}, \\[3pt]
        \mathcal{L}_{\mathrm{Rank}}^{\mathrm{MRL}},
        & \text{for reranker-scored batches}.
    \end{cases}
    \label{eq:stage2_task_objective}
\end{equation}

Reranker-scored batches use the ranking objective in place of the standard contrastive objective. Embedding distillation, by contrast, is added to the selected task objective whenever a larger embedding teacher is available. The overall Stage~2 objective is therefore

\begin{equation}
    \mathcal{L}_{\mathrm{Stage2}}
    =
    \mathcal{L}_{\mathrm{Task}}
    +
    \lambda_{\mathrm{Emb}}
    \mathcal{L}_{\mathrm{Emb}},
    \label{eq:stage2_objective}
\end{equation}

where $\lambda_{\mathrm{Emb}}$ denotes the weight assigned to the embedding-distillation loss.

For the 2B and 4B variants, the frozen 9B WeMM-Embedding model serves as the embedding teacher during Stage~2. Since no larger embedding teacher is available for the 9B variant, we instead train multiple specialized Stage~2 variants using complementary data mixtures and training configurations, and combine them through model merging~\citep{yadav2023ties} to obtain the final 9B model.

\section{Experiments and Analysis}
\label{sec:experiments}

We evaluate WeMM-Embedding across a broad range of public and in-house benchmarks covering image, video, visual-document, text, and agent-oriented retrieval. Our experiments cover the MMEB benchmark series~\cite{vlm2vec-v2,mmeb-v3}, a collection of widely used cross-modal retrieval benchmarks reported in the Gemini Embedding 2 study~\cite{gemini-emb-2}, and an in-house benchmark comprising 26 tasks derived from real-world applications within WeChat. We compare WeMM-Embedding with representative state-of-the-art multimodal embedding models, including both open-source and proprietary commercial models.

\subsection{Performance on the MMEB Series}
\label{subsec:mmeb_series}

The MMEB series provides a comprehensive evaluation across diverse modalities, task formulations, and retrieval scenarios. Specifically, MMEB-v2~\cite{vlm2vec-v2} evaluates general multimodal embedding capabilities across images, videos, and visual documents, while MMEB-v3~\cite{mmeb-v3} further extends the evaluation to audio, complex text retrieval, and agent-centric scenarios. In this section, we follow the benchmark-defined evaluation metrics, using Hit@1 for image, video, audio, and agent tasks and NDCG@5 for text and visual-document tasks.

\subsubsection{General Multimodal Representation}
\label{subsec:general_multimodal_representation}

We first report the performance of WeMM-Embedding on MMEB-v2, which comprises 78 datasets spanning images, videos, and visual documents and covers diverse tasks including classification, question answering, retrieval, and visual grounding~\cite{vlm2vec-v2}. As shown in \Cref{tab:mmeb_v2}, WeMM-Embedding-2B achieves an overall score of 77.9, outperforming Qwen3-VL-Embedding-2B~\cite{qwen3vl-emb} and DME-2B~\cite{DME} by 4.7 and 3.1 points, respectively, and slightly surpassing Qwen3-VL-Embedding-8B. The 4B variant further improves the score to 79.2, outperforming all compared 8B--9B baselines. Scaling to 9B further raises the overall score to 80.6, ranking first on the official MMEB-v2 leaderboard and outperforming all listed open-source and proprietary models.

\begin{table*}[t]
\centering
\scriptsize
\setlength{\tabcolsep}{2.5pt}
\renewcommand{\arraystretch}{1.02}
\resizebox{\textwidth}{!}{%
\begin{tabular}{ll c ccccc ccccc c}
\toprule
\multirow{2}{*}{\textbf{Model}} & \multirow{2}{*}{\textbf{Size}} & \multirow{2}{*}{\textbf{AVG}} & \multicolumn{5}{c}{\textbf{Image}} & \multicolumn{5}{c}{\textbf{Video}} & \textbf{VisDoc} \\
\cmidrule(lr){4-8} \cmidrule(lr){9-13} \cmidrule(lr){14-14}
& & & \textbf{Overall} & \textbf{CLS} & \textbf{QA} & \textbf{Ret} & \textbf{GD} & \textbf{Overall} & \textbf{CLS} & \textbf{QA} & \textbf{V-Ret} & \textbf{M-Ret} & \textbf{Overall} \\
\midrule
\texttt{\# datasets} $\rightarrow$ & & \texttt{78} & \texttt{36} & \texttt{10} & \texttt{10} & \texttt{12} & \texttt{4} & \texttt{18} & \texttt{5} & \texttt{5} & \texttt{5} & \texttt{3} & \texttt{24} \\
\midrule
VLM2Vec~\cite{vlm2vec2024} & 2B & 47.8 & 59.7 & 58.7 & 49.3 & 65.0 & 72.9 & 29.0 & 33.4 & 30.5 & 20.6 & 33.0 & 44.0 \\
GME~\cite{zhang2024gme} & 2B & 55.4 & 51.9 & 54.4 & 29.9 & 66.9 & 55.5 & 33.9 & 34.9 & 42.0 & 25.6 & 32.4 & 76.8 \\
VLM2Vec-V2~\cite{vlm2vec-v2} & 2B & 59.3 & 64.9 & 62.9 & 56.3 & 69.5 & 77.3 & 34.9 & 39.3 & 34.3 & 28.8 & 38.5 & 69.2 \\
Ops-MM-embedding-v1 & 2B & 64.6 & 69.0 & 68.1 & 65.1 & 69.2 & 80.9 & 47.6 & 53.6 & 55.7 & 41.8 & 33.7 & 70.8 \\
RzenEmbed~\cite{jian2025rzenembed} & 2B & 67.2 & 72.3 & 68.5 & 66.3 & 74.5 & 90.3 & 47.3 & 50.4 & 49.7 & 46.6 & 38.9 & 74.5 \\
Qwen3-VL-Embedding~\cite{qwen3vl-emb} & 2B & 73.2 & 75.0 & 70.3 & 74.3 & 74.9 & 88.6 & 61.9 & 71.9 & 64.9 & 53.9 & 53.3 & 79.2 \\
DME-Small~\cite{DME}$^{\dagger}$ & 2B & 74.8 & 75.9 & 68.6 & 76.2 & 75.6 & 94.3 & 65.6 & 84.5 & 61.9 & 55.5 & 57.4 & 79.9 \\
\rowcolor{gray!10}\textbf{WeMM-Embedding} & \textbf{2B} & \underline{77.9} & \underline{79.6} & 74.7 & 79.2 & 79.3 & 94.0 & \underline{70.8} & 84.9 & 71.7 & 63.4 & 58.4 & \underline{80.7} \\
\rowcolor{gray!10}\textbf{WeMM-Embedding} & \textbf{4B} & \textbf{79.2} & \textbf{80.8} & 75.1 & 81.0 & 80.8 & 95.0 & \textbf{72.1} & 86.8 & 72.5 & 65.2 & 58.6 & \textbf{82.0} \\
\midrule
VLM2Vec~\cite{vlm2vec2024} & 8B & 53.2 & 65.5 & 62.7 & 56.9 & 69.4 & 82.2 & 34.0 & 39.1 & 30.0 & 29.0 & 40.6 & 49.1 \\
GME~\cite{zhang2024gme} & 8B & 59.2 & 56.0 & 57.7 & 34.7 & 71.2 & 59.3 & 38.6 & 37.4 & 50.4 & 28.4 & 38.2 & 79.3 \\
Ops-MM-embedding-v1 & 8B & 68.9 & 72.7 & 69.7 & 69.6 & 73.1 & 87.2 & 53.8 & 59.7 & 62.2 & 45.7 & 43.2 & 74.4 \\
RzenEmbed~\cite{jian2025rzenembed} & 8B & 72.9 & 75.9 & 70.6 & 71.7 & 78.5 & 92.1 & 55.7 & 58.8 & 63.5 & 51.0 & 45.5 & 81.3 \\
IFM-TTE~\cite{tte} & 8B & 74.1 & 77.9 & 76.7 & 78.5 & 74.6 & 89.3 & 59.2 & 60.5 & 67.9 & 51.7 & 54.9 & 79.5 \\
Qwen3-VL-Embedding~\cite{qwen3vl-emb} & 8B & 77.8 & \underline{80.1} & 74.2 & 81.1 & 80.2 & 92.3 & 67.1 & 78.4 & 71.0 & 58.7 & 56.1 & \underline{82.4} \\
DME-Medium~\cite{DME}$^{\dagger}$ & 9B & \underline{78.4} & 79.8 & 74.5 & 80.9 & 78.2 & 94.6 & \underline{70.8} & 87.7 & 71.0 & 61.0 & 58.5 & 82.0 \\
\rowcolor{gray!10}\textbf{WeMM-Embedding} & \textbf{9B} & \textbf{80.6} & \textbf{81.9} & 76.2 & 82.2 & 81.7 & 95.6 & \textbf{74.3} & 87.4 & 77.7 & 67.4 & 58.5 & \textbf{83.3} \\
\midrule
Seed-1.6-Embedding$^{\star}$ & -- & 76.9 & 78.0 & 75.1 & 74.9 & 79.3 & 89.1 & 67.7 & 85.2 & 66.7 & 59.1 & 54.8 & 82.2 \\
QQMM-embed-v4$^{\dagger}$ & -- & 78.3 & 82.0 & 76.8 & 83.3 & 80.7 & 96.1 & 67.4 & 77.8 & 71.9 & 60.0 & 54.9 & 80.9 \\
Octen-VL-Large$^{\star}$ & -- & 80.1 & 81.9 & 75.7 & 84.5 & 80.6 & 94.4 & 76.0 & 87.1 & 82.1 & 68.0 & 60.4 & 80.5 \\
DME-Large~\cite{DME}$^{\dagger}$ & -- & 80.2 & 81.1 & 75.4 & 82.9 & 79.6 & 95.2 & 74.4 & 90.1 & 75.7 & 64.5 & 62.2 & 83.4 \\
\bottomrule
\end{tabular}%
}
\caption{Benchmarking results on MMEB-v2. Baseline results are taken from the \href{https://huggingface.co/spaces/TIGER-Lab/MMEB-Leaderboard}{official MMEB leaderboard}. 
$^{\dagger}$Closed-source leaderboard submission without publicly released model weights or a public inference endpoint. $^{\star}$Proprietary commercial model with an undisclosed parameter count. CLS: classification; QA: question answering; Ret: retrieval; GD: visual grounding; V-Ret: video retrieval; M-Ret: moment retrieval.}
\label{tab:mmeb_v2}
\end{table*}

\subsubsection{Text and Agent-Centric Tasks}
\label{subsec:text_agent_retrieval}

We further evaluate WeMM-Embedding on MMEB-v3~\cite{mmeb-v3}, which extends the benchmark with complex text retrieval, agent-centric tasks, audio evaluation, and the MCMR image-retrieval task. Among the newly added tasks, 53 text tasks cover reasoning retrieval, instruction following, long-context retrieval, multi-condition retrieval, and general retrieval, while 47 agent tasks cover tool, GUI, and memory retrieval. When computing V3-All, unsupported tasks are assigned a score of zero; accordingly, the 11 audio tasks are scored as zero for the current WeMM-Embedding models, which do not support audio input. Baseline per-task results on the newly added tasks are obtained from the MMEB-v3 paper or score files submitted to the official MMEB leaderboard.\footnote{Leaderboard results are based on submissions available as of August 24, 2026.} For VLM2Vec~\cite{vlm2vec2024}, VLM2Vec-V2~\cite{vlm2vec-v2}, GME~\cite{zhang2024gme}, and Qwen3-VL-Embedding~\cite{qwen3vl-emb}, we recompute V3-All by combining their originally reported MMEB-v2 results with the corresponding results on the newly added MMEB-v3 tasks.

As shown in \Cref{tab:mmeb_v3}, WeMM-Embedding-2B already outperforms all compared baselines on MMEB-v3, achieving 56.0 on V3-All. It also attains the highest scores among all baselines on the Text and Agent task groups, with 45.3 and 45.1, respectively. The 4B and 9B variants achieve V3-All scores of 58.2 and 59.5, respectively, further widening the gap over existing models.

\begin{table*}[t]
\centering
\scriptsize
\setlength{\tabcolsep}{3.0pt}
\renewcommand{\arraystretch}{1.05}
\resizebox{\textwidth}{!}{%
\begin{tabular}{ll c cccccc cccc cc}
\toprule
\multirow{2}{*}{\textbf{Model}} &
\multirow{2}{*}{\textbf{Size}} &
\multirow{2}{*}{\textbf{V3-All}} &
\multicolumn{6}{c}{\textbf{Text}} &
\multicolumn{4}{c}{\textbf{Agent}} &
\multirow{2}{*}{\textbf{MCMR}} &
\multirow{2}{*}{\textbf{Audio}} \\
\cmidrule(lr){4-9} \cmidrule(lr){10-13}
& & &
\textbf{Overall} & \textbf{RR} & \textbf{IF} & \textbf{LC} & \textbf{MC} & \textbf{GR} &
\textbf{Overall} & \textbf{Tool} & \textbf{GUI} & \textbf{Memory} &
& \\
\midrule
\texttt{\# tasks} $\rightarrow$ & &
\texttt{190} &
\texttt{53} & \texttt{20} & \texttt{9} & \texttt{6} & \texttt{5} & \texttt{13} &
\texttt{47} & \texttt{35} & \texttt{8} & \texttt{4} &
\texttt{1} & \texttt{11} \\
\midrule

VLM2Vec-V2~\cite{vlm2vec-v2}
& 2B & 38.3
& 24.5 & 7.8 & 29.2 & 11.5 & 50.3 & 41.2
& 28.7 & 27.6 & 36.2 & 23.3
& 4.1 & 0.0 \\

Omni-Embed-Nemotron~\cite{omni-embed-nemotron}
& 3B & 43.5
& 39.2 & 17.2 & 42.8 & 40.0 & 69.7 & 56.9
& 36.5 & 38.1 & 32.0 & 32.2
& 26.1 & \textbf{36.5} \\

E5-Omni~\cite{e5-omni}
& 3B & 44.6
& 26.7 & 11.5 & 44.4 & 8.3 & 60.7 & 34.4
& 36.9 & 37.7 & 35.6 & 32.3
& 31.9 & \underline{30.8} \\

Qwen3-VL-Embedding~\cite{qwen3vl-emb}
& 2B & 50.9
& 39.2 & 16.6 & 40.6 & 53.1 & 61.2 & 58.0
& 39.3 & 42.6 & 30.4 & 28.4
& 42.0 & 0.0 \\

\rowcolor{gray!10}
\textbf{WeMM-Embedding}
& \textbf{2B} & \underline{56.0}
& \underline{45.3} & 26.3 & 49.3 & 61.9 & 59.7 & 58.7
& \underline{45.1} & 46.2 & 38.9 & 47.2
& \textbf{42.5} & 0.0 \\

\rowcolor{gray!10}
\textbf{WeMM-Embedding}
& \textbf{4B} & \textbf{58.2}
& \textbf{47.9} & 30.0 & 52.0 & 62.4 & 60.1 & 61.1
& \textbf{49.0} & 50.6 & 41.5 & 50.4
& \underline{41.9} & 0.0 \\

\midrule

WAVE~\cite{wave}
& 7B & 26.3
& 13.7 & 5.9 & 31.3 & 2.6 & 22.9 & 18.6
& 11.3 & 11.9 & 11.8 & 5.7
& 8.9 & 31.8 \\

VLM2Vec~\cite{vlm2vec2024}
& 8B & 32.9
& 22.2 & 7.2 & 28.1 & 5.9 & 40.9 & 41.7
& 19.7 & 19.8 & 21.4 & 15.9
& 0.9 & 0.0 \\

LCO-Embedding-Omni~\cite{lco-embedding}
& 7B & 40.6
& 32.4 & 9.5 & 52.0 & 12.6 & 61.4 & 52.1
& 27.8 & 29.0 & 25.0 & 23.0
& 20.0 & \textbf{43.2} \\

GME~\cite{zhang2024gme}
& 8B & 43.6
& 37.1 & 12.5 & 52.4 & 17.8 & 59.0 & 62.5
& 35.6 & 39.0 & 30.0 & 17.1
& 27.3 & 0.0 \\

E5-Omni~\cite{e5-omni}
& 7B & 47.1
& 26.9 & 11.1 & 42.1 & 9.7 & 56.9 & 35.9
& 36.7 & 37.4 & 38.0 & 27.3
& 41.1 & \underline{43.0} \\

Tianmu-Emb-Uni
& 8B & 53.3
& \underline{43.6} & 20.7 & 47.1 & 58.5 & 63.4 & 62.0
& \underline{39.4} & 42.2 & 35.6 & 22.6
& 38.8 & 38.9 \\

Qwen3-VL-Embedding~\cite{qwen3vl-emb}
& 8B & \underline{53.5}
& 42.5 & 18.2 & 44.8 & 58.0 & 61.2 & 61.2
& 38.4 & 41.3 & 33.5 & 22.8
& 38.0 & 0.0 \\

\rowcolor{gray!10}
\textbf{WeMM-Embedding}
& \textbf{9B} & \textbf{59.5}
& \textbf{48.8} & 31.8 & 52.8 & 64.5 & 58.3 & 61.1
& \textbf{51.0} & 53.0 & 43.3 & 48.9
& \textbf{49.3} & 0.0 \\

\bottomrule
\end{tabular}%
}
\caption{Evaluation results on MMEB-v3. V3-All averages all 190 tasks, comprising the 78 MMEB-v2 tasks, 53 Text tasks, 47 Agent tasks, 11 Audio tasks, and MCMR~\cite{chow2026merit}. Following the evaluation protocol defined in the MMEB-v3 paper~\cite{mmeb-v3}, Text results are reported using NDCG@5. Unsupported tasks are assigned a score of zero. RR: reasoning retrieval; IF: instruction following; LC: long-context retrieval; MC: multi-condition retrieval; GR: general retrieval.}
\label{tab:mmeb_v3}
\end{table*}

\subsection{Cross-Modal Retrieval Evaluation}
\label{subsec:cross_modal_retrieval}

The cross-modal evaluation comprises 12 widely used public benchmarks that measure semantic alignment between text and visual content across images, videos, and visual documents. For the three proprietary models, Gemini Embedding 2~\cite{gemini-emb-2}, Amazon Nova MME~\cite{amazon-nova-mme}, and Voyage Multimodal 3.5~\cite{voyage3.5}, we use the results reported in the Gemini Embedding 2 technical report~\cite{gemini-emb-2}. Qwen3-VL-Embedding and WeMM-Embedding are evaluated by us on the corresponding public evaluation sets using the metric specified for each benchmark.

As shown in \Cref{tab:cross_modal_retrieval}, WeMM-Embedding-2B achieves an overall score of 79.8 across the 12 tasks, outperforming all compared open-source baselines and comparing favorably with leading proprietary models. Scaling the model to 4B and 9B further improves the overall score to 80.8 and 81.7, respectively.

\begingroup
\newcommand{\modelhead}[2]{\shortstack[c]{\strut\textbf{#1}\\\strut\textbf{#2}}}
\newcommand{\closedhead}[2]{\shortstack[c]{\strut\textbf{#1}\\\strut\textbf{#2}\rlap{\smash{$^{\dagger}$}}}}

\begin{table*}[htbp]
\centering
\scriptsize
\setlength{\tabcolsep}{4.0pt}
\renewcommand{\arraystretch}{1.06}
\resizebox{\textwidth}{!}{%
\begin{tabular}{ll cccc >{\columncolor{gray!10}}c >{\columncolor{gray!10}}c >{\columncolor{gray!10}}c}
\toprule
& & \closedhead{Gemini}{Embedding 2} & \closedhead{Amazon}{Nova MME} & \closedhead{Voyage}{Multimodal 3.5} & \modelhead{Qwen3-VL-}{Embedding} & \modelhead{WeMM-}{Embedding} & \modelhead{WeMM-}{Embedding} & \modelhead{WeMM-}{Embedding} \\
\midrule
\textbf{Size} & & -- & -- & -- & 8B & 2B & 4B & 9B \\
\midrule
\multirow{5}{*}{\shortstack[l]{\textbf{Text $\rightarrow$ Image}\\Recall@1}} & MSCOCO~\cite{mscoco-chen2015microsoft} & 62.9 & 57.2 & 58.1 & 58.7 & 62.6 & 63.5 & 64.6 \\
& Flickr30k~\cite{plummer2015flickr30k} & 89.1 & 81.6 & 89.9 & 84.6 & 86.9 & 87.0 & 87.8 \\
& DOCCI~\cite{onoe2024docci} & 93.4 & 84.0 & 83.8 & 95.0 & 94.0 & 95.8 & 96.7 \\
& TextCaps~\cite{sidorov2020textcaps} & 89.6 & 76.0 & 79.4 & 79.3 & 82.5 & 85.8 & 85.2 \\
\cmidrule(l){2-9}
& \textbf{Mean} & \textbf{83.8} & 74.7 & 77.8 & 79.4 & 81.5 & 83.0 & \underline{83.6} \\
\midrule
\multirow{5}{*}{\shortstack[l]{\textbf{Image $\rightarrow$ Text}\\Recall@1}} & MSCOCO~\cite{mscoco-chen2015microsoft} & 78.8 & 68.3 & 74.5 & 71.1 & 78.8 & 79.4 & 79.3 \\
& Flickr30k~\cite{plummer2015flickr30k} & 97.4 & 87.5 & 94.5 & 90.7 & 95.6 & 96.6 & 94.4 \\
& DOCCI~\cite{onoe2024docci} & 91.3 & 76.5 & 77.4 & 92.8 & 90.6 & 93.7 & 92.8 \\
& TextCaps~\cite{sidorov2020textcaps} & 97.4 & 88.9 & 88.6 & 91.9 & 93.3 & 96.0 & 94.8 \\
\cmidrule(l){2-9}
& \textbf{Mean} & 91.2 & 80.3 & 83.8 & 86.6 & 89.6 & \underline{91.4} & \textbf{90.3} \\
\midrule
\multirow{4}{*}{\shortstack[l]{\textbf{Text $\rightarrow$ Video}\\NDCG@10}} & VATEX~\cite{wang2019vatex} & 68.8 & 60.3 & 55.2 & 68.5 & 74.6 & 75.9 & 78.4 \\
& MSR-VTT~\cite{msr-vtt} & 68.0 & 67.0 & 63.0 & 70.6 & 72.7 & 71.7 & 74.1 \\
& YouCook2~\cite{youcook2} & 52.5 & 34.7 & 31.4 & 54.7 & 63.9 & 62.2 & 66.3 \\
\cmidrule(l){2-9}
& \textbf{Mean} & 63.1 & 54.0 & 49.9 & 64.6 & \underline{70.4} & 69.9 & \textbf{72.9} \\
\midrule
\shortstack[l]{\textbf{Document Retrieval}\\NDCG@10} & ViDoRe V2~\cite{vidore-v2} & {64.9} & 60.6 & \underline{65.5} & 62.8 & 61.4 & 62.1 & \textbf{66.3} \\
\midrule
\textbf{AVG} & & 79.5 & 70.2 & 71.8 & 76.7 & 79.8 & \underline{80.8} & \textbf{81.7} \\
\bottomrule
\end{tabular}%
}
\caption{Cross-modal retrieval results on 12 public benchmarks. AVG denotes the average across the 12 datasets. $^{\dagger}$Proprietary commercial model with an undisclosed parameter count.}
\label{tab:cross_modal_retrieval}
\end{table*}
\endgroup

\subsection{In-House Evaluation}
\label{subsec:business_evaluation}

We further evaluate WeMM-Embedding on an in-house benchmark comprising 26 tasks derived from real-world applications within WeChat. The benchmark covers five categories: classification, search, cross-domain content matching, article relevance, and video relevance. As shown in \Cref{tab:business}, WeMM-Embedding-2B substantially outperforms the representative open-source baseline and achieves higher scores across all five categories.

\begin{table*}[htbp]
\centering
\small
\setlength{\tabcolsep}{5.7pt}
\renewcommand{\arraystretch}{1.06}
\resizebox{\textwidth}{!}{%
\begin{tabular}{ll cccccc}
\toprule
\textbf{Model} & \textbf{Size} & \textbf{AVG} & \textbf{Classification} & \textbf{Search} & \textbf{Cross-DM} & \textbf{Article Rel.} & \textbf{Video Rel.} \\
\midrule
\texttt{\# tasks} $\rightarrow$ & & \texttt{26} & \texttt{4} & \texttt{7} & \texttt{4} & \texttt{4} & \texttt{7} \\
\midrule
Qwen3-VL-Embedding~\cite{qwen3vl-emb} & 2B & 60.9 & 60.1 & 51.9 & 55.5 & 75.5 & 64.9 \\
\rowcolor{gray!10}\textbf{WeMM-Embedding} & \textbf{2B} & \textbf{72.0} & \textbf{72.6} & \textbf{65.5} & \textbf{73.7} & \textbf{86.4} & \textbf{68.8} \\
\bottomrule
\end{tabular}%
}
\caption{Evaluation results on the in-house benchmark. AVG denotes the average over all 26 tasks. Cross-DM denotes cross-domain content matching; Article Rel. denotes article relevance; Video Rel. denotes video relevance.}
\label{tab:business}
\end{table*}

Beyond offline evaluation, WeMM-Embedding has been deployed at scale in recommendation systems spanning WeChat Official Accounts, WeChat Channels, and e-commerce content. Its multimodal representations combine complementary signals from text, cover images, and video frames, and are used at multiple stages of the recommendation pipeline, including candidate retrieval, ranking feature construction, user sequence modeling, and cross-domain content understanding. Semantic IDs derived from these representations further provide compact discrete features for indexing and sequence modeling. To date, WeMM-Embedding has delivered consistent gains in 14 online A/B tests across these systems, with the corresponding improvements subsequently rolled out in production. These deployments have improved content matching, recommendation quality, and user engagement, with notable gains for long-tail and newly published content.

WeMM-Embedding has also been deployed in WeChat search, supporting semantic retrieval across diverse content sources including WeChat Channels videos, WeChat Official Accounts articles, and WeChat Moments. The model supports both unimodal and cross-modal retrieval and improves semantic relevance and retrieval quality across text, image, and video content. Taken together, the in-house evaluation and production deployments show that the strong performance of WeMM-Embedding extends beyond public benchmarks to diverse real-world tasks and large-scale recommendation and search systems.

\subsection{Further Analysis}
\label{subsec:further_analysis}

In this section, we first analyze the performance of the learned Matryoshka representations across different embedding dimensions, then revisit key Stage-1 design choices, and finally present a cumulative analysis of the strategies adopted for Stage-2 training.

\subsubsection{MRL Analysis}
\label{subsec:mrl_analysis}

WeMM-Embedding incorporates Matryoshka Representation Learning (MRL)~\cite{MRL2022}, allowing a single model to produce nested representations at multiple embedding dimensions. We evaluate the 2B model at dimensions ranging from 64 to 2,048 on the corresponding task subsets of MMEB-v2 to examine how dimensionality reduction affects different modalities and task types. For each evaluation group, we report the proportion of performance retained relative to its result at 2,048 dimensions.

\begin{figure*}[htbp]
\centering
\includegraphics[width=0.92\textwidth]{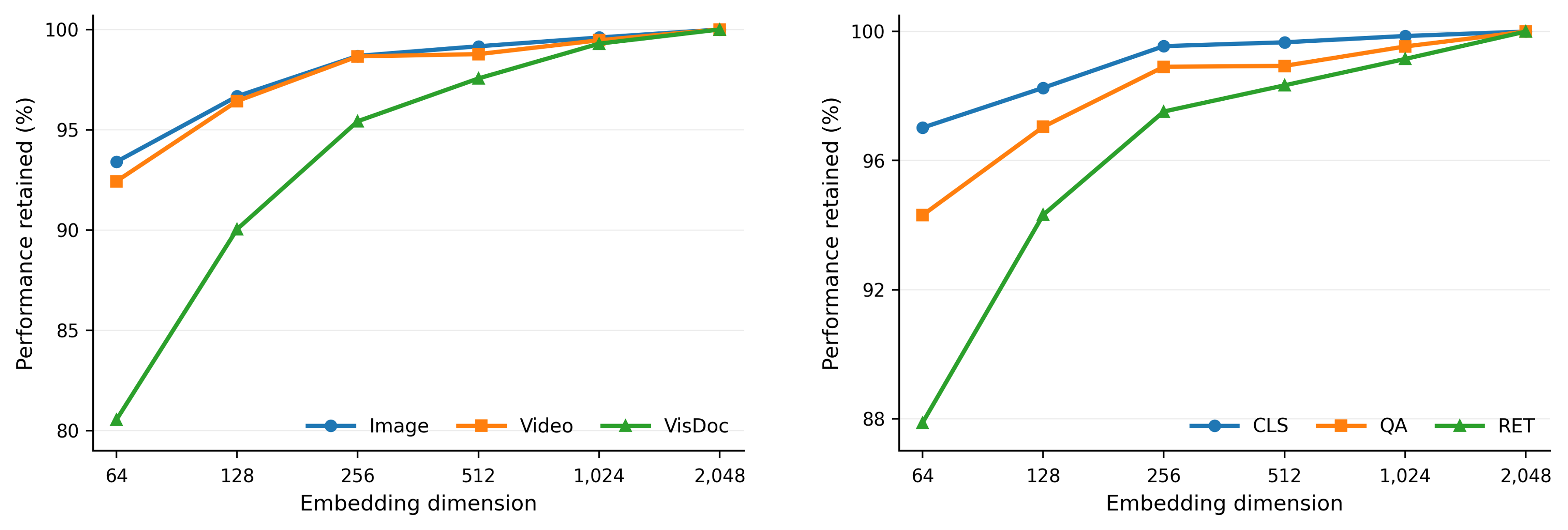}
\caption{\textbf{MRL analysis of WeMM-Embedding-2B on MMEB-v2.} \textbf{Left:} Performance retained on the image, video, and visual-document subsets across embedding dimensions. \textbf{Right:} Performance retained for classification (CLS), question answering (QA), and retrieval (RET), each averaged over the corresponding image and video tasks.}
\label{fig:mrl_analysis}
\end{figure*}

As shown in the left panel of \Cref{fig:mrl_analysis}, performance on image and video tasks follows closely matched trends as the embedding dimension decreases. At 256 dimensions, the model retains 98.7\% of its 2,048-dimensional performance on both image and video tasks; the corresponding retention rates rise to 99.2\% and 98.8\% at 512 dimensions, respectively. Visual-document tasks exhibit greater sensitivity to dimensionality reduction, which may reflect the higher information density of visual documents, particularly their dense textual content.

The right panel of \Cref{fig:mrl_analysis} further shows different sensitivity patterns across task types. Classification is the least sensitive to dimensionality reduction, question answering exhibits moderate degradation, and retrieval is affected most substantially, especially at 64 and 128 dimensions. Once the embedding dimension reaches 256, however, all three task groups retain more than 97\% of their respective full-dimensional performance, and the gains from further increasing the dimension become progressively smaller. Taken together, these results suggest that 256- or 512-dimensional representations provide practical choices for efficiency-sensitive applications while preserving most of the performance achieved at 2,048 dimensions.

\subsubsection{Stage-1 Design Analysis}
\label{subsec:stage1_ablation}

We conduct a small-scale ablation study with a 2B model to examine some of the key Stage-1 design choices, including task-specific instructions, task-consistent batching, and duplicate-aware masking. The resulting variants are evaluated on MMEB-v2 under a common experimental protocol.

\begin{table}[htbp]
\centering
\small
\setlength{\tabcolsep}{5.5pt}
\renewcommand{\arraystretch}{1.06}
\begin{tabular}{lcccc}
\toprule
\textbf{Stage-1 Configuration} & \textbf{AVG} & \textbf{Image} & \textbf{Video} & \textbf{VisDoc} \\
\midrule
\textbf{Full configuration} & \textbf{71.9} & \textbf{76.1} & \textbf{59.1} & \textbf{75.3} \\
\quad \textit{w/o} task-specific instructions & 71.1 & 75.7 & 58.2 & 73.8 \\
\quad \textit{w/o} task-consistent batching & 68.5 & 71.5 & 56.3 & 73.1 \\
\quad \textit{w/o} duplicate-aware masking & 71.4 & 75.3 & 58.6 & 75.1 \\
\bottomrule
\end{tabular}
\caption{Results on MMEB-v2 from a small-scale Stage-1 ablation study using a 2B model. AVG denotes the average across all 78 MMEB-v2 tasks.}
\label{tab:stage1_ablation}
\end{table}

As shown in \Cref{tab:stage1_ablation}, removing task-specific instructions leads to a 0.8-point decrease in the overall score, with the largest degradation observed on visual-document tasks. This result indicates that explicit instructions help the model capture task-specific matching objectives across heterogeneous tasks. Task-consistent batching has the largest impact among the evaluated designs. Replacing it with mixed sampling results in a 3.4-point decrease in overall performance, suggesting that batches constructed from a consistent task and candidate space provide more informative in-batch negatives. Duplicate-aware masking also contributes to performance, as its removal leads to a 0.5-point decrease in the overall score. By filtering duplicate or semantically equivalent targets from the negative set, it makes the unified contrastive objective better suited to tasks with repeated labels, such as classification.

\subsubsection{Cumulative Stage-2 Analysis}
\label{subsec:stage2_ablation}

We further conduct a cumulative study with the 2B model to examine several key strategies used in Stage-2 training, including curated data, reranker supervision, embedding-teacher distillation, and an expanded visual input budget. Starting from the Stage-1 checkpoint, these strategies are introduced sequentially, and each resulting configuration is evaluated on MMEB-v2.

\begin{table}[htbp]
\centering
\small
\setlength{\tabcolsep}{5.5pt}
\renewcommand{\arraystretch}{1.06}
\begin{tabular}{lcccc}
\toprule
\textbf{Cumulative Configuration} & \textbf{AVG} & \textbf{Image} & \textbf{Video} & \textbf{VisDoc} \\
\midrule
Stage-1 checkpoint & 75.7 & 77.7 & 67.5 & 78.9 \\
\quad + curated data & 76.6 & 78.9 & 69.2 & 78.5 \\
\quad + reranker supervision & 76.7 & 78.8 & 69.4 & 79.1 \\
\quad + embedding-teacher distillation & 77.6 & 79.4 & 70.0 & 80.4 \\
\rowcolor{gray!10}\quad \textbf{+ expanded visual input budget (final)} & \textbf{77.9} & \textbf{79.6} & \textbf{70.8} & \textbf{80.7} \\
\bottomrule
\end{tabular}
\caption{Cumulative Stage-2 results of the 2B model on MMEB-v2. Each row reports the configuration after introducing the listed strategy. AVG denotes the average across all 78 MMEB-v2 tasks.}
\label{tab:stage2_ablation}
\end{table}

As shown in \Cref{tab:stage2_ablation}, cumulatively incorporating our Stage-2 strategies improves the overall score by 2.2 points. First, curated data yields clear gains on both image and video tasks, highlighting the value of balanced resampling, quality refinement, and hard-negative enrichment when constructing a compact Stage-2 training set from the large-scale multimodal corpus. Building on this, reranker supervision provides an additional improvement, with its largest benefit observed on visual-document tasks. Embedding-teacher distillation further improves performance across all three domains, demonstrating that dense similarity supervision transfers effectively across heterogeneous tasks and contributes to the strong performance of the compact model. Finally, expanding the visual input budget through higher-resolution inputs and denser video frame sampling brings a further improvement, particularly on video tasks. Overall, these substantial gains suggest that carefully designed supervision signals remain essential for further refining universal multimodal embedding models once large-scale multimodal alignment has been established.

\section{Conclusion and Future Work}
\label{sec:conclusion}

In this report, we introduced WeMM-Embedding, a family of universal multimodal embedding models spanning multiple scales. The family is trained with a two-stage strategy that progresses from broad multimodal alignment to finer-grained representation learning through curated data, richer relevance supervision, and cross-scale knowledge transfer. Extensive evaluations across public benchmarks demonstrate that WeMM-Embedding achieves state-of-the-art performance and establishes a new performance--efficiency frontier. Moreover, it delivers consistent gains on the 26-task in-house benchmark and across 14 online A/B tests. WeMM-Embedding has also been deployed at scale across multiple recommendation and search systems within WeChat, further demonstrating its practical value. In the future, we will continue to extend WeMM-Embedding toward omni-modal inputs, scale the model family to larger sizes, and further improve data curation and fine-grained relevance supervision.

\bibliography{main}
\bibliographystyle{plainnat}

\end{document}